\documentclass[10pt,twocolumn,letterpaper]{article}

\usepackage[letterpaper,
            top=1.0in, bottom=1.125in,
            left=0.875in, right=0.875in,
            columnsep=0.3125in]{geometry}

\usepackage{times}
\usepackage[T1]{fontenc}
\usepackage[utf8]{inputenc}
\usepackage{microtype}

\usepackage{amsmath,amssymb,amsfonts}

\usepackage{graphicx}
\usepackage{booktabs}
\usepackage[font=small,labelfont=bf,labelsep=period]{caption}

\usepackage[numbers,sort&compress]{natbib}

\usepackage[breaklinks=true,bookmarks=false,hidelinks]{hyperref}

\usepackage{titlesec}
\titleformat{\section}{\normalfont\large\bfseries}{\thesection.}{0.5em}{}
\titleformat{\subsection}{\normalfont\normalsize\bfseries}{\thesubsection.}{0.5em}{}
\titlespacing*{\section}{0pt}{1.5ex plus 0.5ex minus 0.2ex}{1ex plus 0.2ex}
\titlespacing*{\subsection}{0pt}{1.2ex plus 0.4ex minus 0.2ex}{0.8ex plus 0.2ex}

\makeatletter
\renewcommand{\@maketitle}{%
  \newpage
  \null
  \vskip 1em%
  \begin{center}%
    \let\footnote\thanks
    {\LARGE\bfseries \@title \par}%
    \vskip 1.2em%
    {\large \@author \par}%
    \vskip 0.5em%
    {\normalsize \@date \par}%
  \end{center}%
  \par
  \vskip 1em}
\makeatother

\newcommand{\R}{\mathbb{R}}
\title{\textbf{GC-ART}: Global Learnable Second-Order Rational Tone Curves for Illumination Robustness}

\author{%
  Wei Huang\textsuperscript{1,*}
  \qquad
  Joyce Huang\textsuperscript{2,*}
  \\[0.5em]
  \textsuperscript{1}\textit{Microsoft}
  \qquad
  \textsuperscript{2}\textit{Massachusetts Institute of Technology}
  \\[0.5em]
  \textsuperscript{*}Equal contribution.
  \\[0.75em]
  {\small The views expressed in this paper are those of the authors and do not necessarily reflect the views of their employer or institution.}
}
\date{}

\renewenvironment{abstract}
  {\centerline{\large\bfseries Abstract}\vspace{0.5em}\begin{quote}\small}
  {\end{quote}\vspace{0.5em}}

\begin{document}

\twocolumn[{%
\renewcommand\twocolumn[1][]{#1}%
\maketitle

\begin{abstract}
We introduce \textbf{GC-ART} (Global Curve Adaptive Rational Tone-mapping), a lightweight differentiable pre-processing module for robust image classification. GC-ART predicts an endpoint-pinned rational tone curve from per-channel soft histograms using a 643-parameter MLP, then applies the curve pointwise before the classifier. The module is trained end-to-end with cross-entropy and a soft monotonicity penalty. On CIFAR-10 with a CIFAR-style ResNet-18, GC-ART matches clean accuracy with the unenhanced baseline and other learned enhancers, improves over the baseline on multiplicative darkening, and achieves the best learned-method result on contrast corruption (48.45\% vs. 46.27\% for the baseline and 47.13\% for Zero-DCE++). These results suggest that histogram-conditioned rational curves can learn useful global tone corrections, including contrast-expanding behavior, while preserving edge locations by construction through pointwise mapping. GC-ART also uses substantially fewer FLOPs than convolutional learned enhancers at $32\times32$. The current hyperparameters are untuned, leaving room for systematic improvement.
\end{abstract}
\vspace{1em}
}]


\section{Introduction}
Image classifiers often degrade when the illumination statistics at test time differ from those seen during training~\cite{hendrycks2019robustness}. A natural response is to insert an image-enhancement module before the classifier and train the composite system end-to-end. Existing enhancement networks, however, were largely developed for perceptual image restoration rather than downstream recognition. They typically use spatial convolutional architectures and are trained with perceptual or reference-based losses designed to produce visually pleasing images~\cite{guo2020zerodce, ma2022sci, wei2018retinex, zhang2019kind}.

Recent learned tone-mapping methods can be organized into several architectural families. Bilateral-grid methods such as HDRNet~\cite{gharbi2017deep} predict spatially varying affine coefficients on a low-resolution grid and slice the grid to recover full-resolution outputs. Image-adaptive 3D-LUT methods~\cite{zeng2020learning} predict per-image blending weights over a learned basis of lookup tables. Iterative-polynomial methods, including Zero-DCE~\cite{guo2020zerodce} and related systems~\cite{ma2022sci}, predict pixel-wise coefficients for a fixed quadratic light-enhancement curve, $f(x)=x+\alpha x(1-x)$, and apply it repeatedly. Direct image-to-image methods, including U-Net-style low-light enhancers~\cite{chen2018learning, lore2017llnet, wei2018retinex, zhang2019kind}, avoid an explicit curve and predict enhanced pixels directly. In contrast, rational functions have long been used in classical graphics tone mapping, including Reinhard, Filmic, and ACES-style curves~\cite{reinhard2002photographic,hable2010uncharted,narkowicz2016aces}, but they have not commonly been used as learned, per-image tone curves for classifiers.

This paper studies whether such a curve family is useful as a recognition front-end. We propose GC-ART, a global per-channel tone-mapping module whose parameters are predicted from a per-image soft histogram by a 643-parameter MLP. The learned component has resolution-independent cost, while the final curve application is pointwise. This design has two intended advantages. First, the rational form can represent, within a single parameterization, both concave shadow-lifting and convex highlight-compressing shapes (Section~\ref{sec:curve}). Second, because the module only remaps pixel intensities using image-global curve parameters, it cannot spatially average across an edge; unlike local image-to-image convolutional enhancers, it can change contrast without moving or smoothing edge locations by construction.

The resulting evidence is encouraging but bounded. On CIFAR-10 with synthetic illumination corruptions, GC-ART maintains the clean accuracy of the unenhanced classifier and the Zero-DCE-style learned enhancers, improves over the unenhanced baseline on multiplicative darkening, and achieves the strongest learned-method performance on contrast corruption. The darkening result shows that the module can provide a useful learned correction for underexposure; the contrast result is even more directly aligned with the bidirectional tone-shaping motivation, because recovering contrast requires a curve that expands compressed intensity ranges rather than only lifting shadows or compressing highlights. The result therefore supports GC-ART as a compact global front-end for recognition, with two attractive properties: image-conditioned rational curves can express multiple correction directions in a single module, and pointwise curve application preserves edge locations rather than performing local spatial filtering. Classical Histogram Equalization remains stronger on this particular synthetic contrast benchmark, making it an important reference point and motivating future work on combining GC-ART's learnability with the robustness of stronger histogram-based corrections.


\section{Related Work}

\textbf{Classical and graphics-pipeline tone mapping.}
Tone mapping in computer graphics maps high-dynamic-range luminance to a displayable range. Reinhard et al.~\cite{reinhard2002photographic} introduced the rational global operator $L_d=L/(1+L)$ for HDR-to-display compression, and Drago et al.~\cite{drago2003adaptive} proposed an adaptive logarithmic mapping for high-contrast scenes. Real-time rendering pipelines often use higher-order rational curves, including the Filmic / Uncharted~2 curve of Hable~\cite{hable2010uncharted} and ACES-style fitted curves~\cite{narkowicz2016aces}. TinyML-based Global Tone Mapping (TGTM)~\cite{todorov2024tgtm} also predicts global tone-mapping behavior from histogram information for efficient HDR-sensor tone mapping. Its goal, however, is HDR-to-display image reconstruction and perceptual image quality, whereas GC-ART is trained as a differentiable recognition front-end. TGTM is therefore not studied as a fully differentiable module jointly optimized with a downstream classifier; GC-ART is optimized end-to-end for machine-vision accuracy.

\textbf{Learned image enhancement.}
Deep enhancement methods include autoencoder-based approaches such as LLNet~\cite{lore2017llnet}, Retinex-style decompositions such as RetinexNet and KinD~\cite{wei2018retinex,zhang2019kind}, curve-estimation methods such as Zero-DCE~\cite{guo2020zerodce} and SCI~\cite{ma2022sci}, differentiable image-processing pipelines such as Exposure~\cite{hu2018exposure}, image-adaptive 3D-LUTs~\cite{zeng2020learning}, bilateral-grid methods such as HDRNet~\cite{gharbi2017deep}, and direct image-to-image architectures~\cite{chen2018learning}. These methods are primarily designed for perceptual enhancement and are commonly trained with perceptual, reference, or zero-reference losses. Our comparison in Section~\ref{sec:setup} instead isolates a classification setting: all learned front-ends are trained only through downstream cross-entropy. The Zero-DCE-style baselines are therefore architectural simplifications, not evaluations of the original Zero-DCE method or training recipe.

\textbf{Histogram-based and histogram-conditioned methods.}
Classical Histogram Equalization~\cite{gonzalez2018dip} maps intensities through the empirical CDF, and CLAHE~\cite{zuiderveld1994clahe} applies contrast-limited local equalization. Both are fixed pre-processors rather than trainable modules. Differentiable histograms based on soft binning have also been used as loss terms or feature extractors in deep learning~\cite{ustinova2016learning,avi2020deephist}. GC-ART uses a Gaussian-RBF soft histogram as the input to a small parameter-prediction Hyper-Network~\cite{ha2017hypernetworks}, giving the histogram a direct architectural role in choosing the tone curve.


\section{Method}
GC-ART consists of three differentiable steps: soft histogram extraction, histogram-conditioned parameter prediction, and pointwise application of a rational tone curve. During training, we add a soft monotonicity penalty to discourage locally decreasing curves.

\subsection{Differentiable Soft Histogram}
Let $\mathbf{X}\in\R^{H\times W\times C}$ denote an image with intensities in $[0,1]$, and let $\{c_i\}_{i=1}^{K}$ be uniformly spaced bin centers in $[0,1]$. For each channel $c$, GC-ART computes a soft histogram $\mathbf{h}_c\in\R^K$ by averaging Gaussian RBF responses over spatial locations:
\begin{equation}
h_{c,i} \;=\; \frac{1}{HW}\sum_{u,v}\exp\!\left(-\frac{(X_{u,v,c}-c_i)^2}{\gamma}\right).
\label{eq:hist}
\end{equation}
The histogram is computed separately for each image and channel; no dataset-level statistics are used. In all experiments, $K=16$ and $\gamma=0.01$.

\subsection{Rational Tone-Mapping Curve}
\label{sec:curve}
For each channel, GC-ART applies a rational tone curve on $[0,1]$:
\begin{equation}
f(x;\,a,b,d,e) \;=\; \frac{a\,x^{2} + b\,x}{d\,x^{2} + e\,x + 1}.
\label{eq:curve}
\end{equation}
The missing constant term in the numerator ensures $f(0)=0$. We enforce the upper endpoint by setting
\begin{equation}
b \;=\; d + e + 1 - a,
\label{eq:bconstraint}
\end{equation}
which gives $f(1)=1$. A Hyper-Network $\mathcal{H}:\R^K\to\R^3$ predicts $(a,\tilde d,\tilde e)$ for each channel from its soft histogram. We apply softplus to $\tilde d$ and $\tilde e$ to obtain $d,e>0$, making the denominator at least one on $[0,1]$ and preventing singularities in the input range.

\textbf{Range.}
The endpoints are fixed, but the implementation does not clamp intermediate values. For some parameter choices, $f(x)$ can therefore lie outside $[0,1]$ for intermediate $x$.

\textbf{Motivation.}
The rational family is intended to provide a compact curve class that can represent multiple exposure corrections with one parameterization. Depending on $(a,d,e)$, Equation~\ref{eq:curve} can approximate concave shadow-lifting curves, convex highlight-compressing curves, sigmoidal curves that trade shadow and highlight detail for mid-tone contrast, and near-identity curves. Fixed gamma correction lacks this per-image bidirectionality: a chosen $\gamma$ favors one correction direction and is shared by all images.

\textbf{Connection to graphics tone mapping.}
Rational tone curves are standard in graphics pipelines. Reinhard's operator~\cite{reinhard2002photographic}, $L_d=L/(1+L)$, is a degree-1 rational; Filmic / Uncharted~2~\cite{hable2010uncharted} and ACES-style curves~\cite{narkowicz2016aces} use related higher-order forms. Those curves are fixed and human-designed. GC-ART instead predicts curve parameters per image from the input histogram and optimizes them through the downstream classifier. This makes the rational family task-adaptive: the same compact module can in principle choose shadow-lifting, highlight-compressing, or contrast-expanding behavior depending on the image statistics. In our experiments, its clearest learned-method advantage is on contrast corruption, where a single curve must expand compressed low and high intensity ranges. Detailed analysis of the learned curve shapes is left to future work (Section~\ref{sec:future}).

\textbf{Hyper-Network.}
$\mathcal{H}$ is a two-layer MLP with hidden width 32 (16$\,\to\,$32$\,\to\,$3, ReLU~\cite{nair2010relu}), for 643 parameters. The final layer is initialized with zero weights and bias $(0,-5,-5)$. Thus $a=0$ and $d,e=\mathrm{softplus}(-5)\approx6.7\times10^{-3}$ at initialization, making the initial curve within roughly $1\%$ of the identity on $[0,1]$.

\subsection{Soft Monotonicity Penalty}
A non-monotone tone curve can invert intensity order and discard information useful to the classifier. We therefore penalize the negative part of the curve derivative,
\begin{equation}
\mathcal{L}_{\text{mono}} \;=\; \int_0^1 \max\!\left(0, -f'(t)\right)\,dt.
\label{eq:mono_continuous}
\end{equation}
In the implementation, this continuous penalty is estimated on a uniform grid $t_1,\ldots,t_M\in[0,1]$ by penalizing negative adjacent finite differences:
\begin{equation}
\widehat{\mathcal{L}}_{\text{mono}} \;=\; \frac{1}{M-1}\sum_{j=1}^{M-1}\max\!\left(0,\; -\bigl(f(t_{j+1})-f(t_j)\bigr)\right).
\label{eq:mono}
\end{equation}
We use $M=32$, separate from the histogram bin count $K=16$. The training objective is $\mathcal{L}=\mathcal{L}_{\text{CE}}+\lambda\widehat{\mathcal{L}}_{\text{mono}}$ with $\lambda=10$. The estimated penalty is averaged over batch, channel, and finite-difference dimensions before being added to cross-entropy. It is a soft regularizer: it discourages local decreases on the sampled grid but does not guarantee global monotonicity. The values of $\lambda$ and $M$ were chosen manually rather than by a sweep.

\subsection{Why Not Use the CDF Directly?}
A natural alternative is the empirical CDF used in Histogram Equalization. We use a learned parametric curve instead for three reasons. First, the empirical CDF is based on counting and requires a surrogate to provide smooth gradients. Second, CDF mapping remaps intensities toward a more uniform pixel distribution; in extreme low-light OOD images where dark pixels dominate, the CDF curve can exhibit steep gradients that strongly stretch shadow values, potentially amplifying sensor noise and altering features useful for recognition. Third, CDF mapping is a static, task-agnostic formulation designed for general contrast enhancement rather than learned directly for machine vision. GC-ART addresses the first two issues by using a differentiable histogram and learning a smooth curve from the classification objective. We do not compare against a differentiable HE baseline, which remains useful future work.

\section{Experiments}

\subsection{Setup}
\label{sec:setup}
\textbf{Task and corruptions.}
We train on CIFAR-10~\cite{krizhevsky2009cifar} with random brightness jitter in $[0.5,1.0]$ (darkening only), random crop with padding 4, and horizontal flip. Pixel values remain in $[0,1]$ throughout, and we do not apply per-channel mean/std normalization because GC-ART's curve domain is $[0,1]$. We evaluate on the clean test set and on three CIFAR-10-C-style illumination corruptions computed on the fly following Hendrycks and Dietterich~\cite{hendrycks2019robustness}: \textsc{brightness} adds constants $\{0.1,0.2,0.3,0.4,0.5\}$ to the HSV value channel and converts back to RGB, making it an overexposure corruption; \textsc{contrast} multiplies $(x-\bar{x})$ by $\{0.4,0.3,0.2,0.1,0.05\}$ around the per-channel spatial mean; and \textsc{darken} multiplies intensities by $\{0.8,0.6,0.4,0.25,0.1\}$. Thus \textsc{brightness} and \textsc{darken} test opposite directions of illumination shift, while \textsc{contrast} tests dynamic-range collapse.

\textbf{Models.}
All learned systems use the same CIFAR-style ResNet-18~\cite{he2016resnet}: a $3\times3$ stem, no max-pool, and a 10-way classification head. We evaluate four learned front-ends: no enhancer (\textsc{baseline}); GC-ART; a stripped Zero-DCE-style~\cite{guo2020zerodce} convolutional enhancer with 3 convolutional layers, one 3-channel $\alpha$ map, 3 curve iterations, and $\sim$11k parameters; and a smaller depthwise-separable variant denoted Zero-DCE++ with $\sim$2k parameters. Both Zero-DCE-style baselines are trained only with classification cross-entropy and are not the original Zero-DCE architecture or training recipe. We additionally evaluate five fixed classical pre-processors applied at both train and test time: Histogram Equalization (\textsc{he}), CLAHE, and gamma correction with $\gamma\in\{1.5,2.2,3.0\}$.

\textbf{Training.}
All learned models are trained from scratch for 100 epochs with Adam~\cite{kingma2015adam} ($\mathrm{lr}=10^{-3}$), cosine annealing~\cite{loshchilov2017sgdr}, batch size 1024, and bfloat16 mixed precision on a single A100 GPU. Each configuration is trained with seeds 42, 43, and 44. We report the mean and standard deviation across seeds at the final epoch. Each seed is a separate run, so the mini-batch ordering confound present in an earlier pilot is not present in these results.

\textbf{Zero-DCE-style baselines.}
The baselines labeled ``Zero-DCE'' and ``Zero-DCE++'' are intentionally simplified. This choice serves two purposes. First, it reduces the capacity gap: the original Zero-DCE network has roughly 80{,}000 parameters~\cite{guo2020zerodce}, whereas GC-ART has 643 parameters. Our MiniZeroDCE ($\sim$11k) and MiniZeroDCE++ ($\sim$2k) preserve the spatial-convolutional, per-pixel-curve structure while making the comparison less dominated by parameter count. Second, it aligns the objective: original Zero-DCE is trained with perceptual zero-reference losses, whereas our question concerns front-ends optimized only through classification loss. Consequently, our results should be read as comparisons among architectural front-ends under classification-only training, not as claims about the original Zero-DCE method or its perceptual training recipe.

\subsection{Main Results}
\begin{figure*}[t]
\centering
\includegraphics[width=\textwidth]{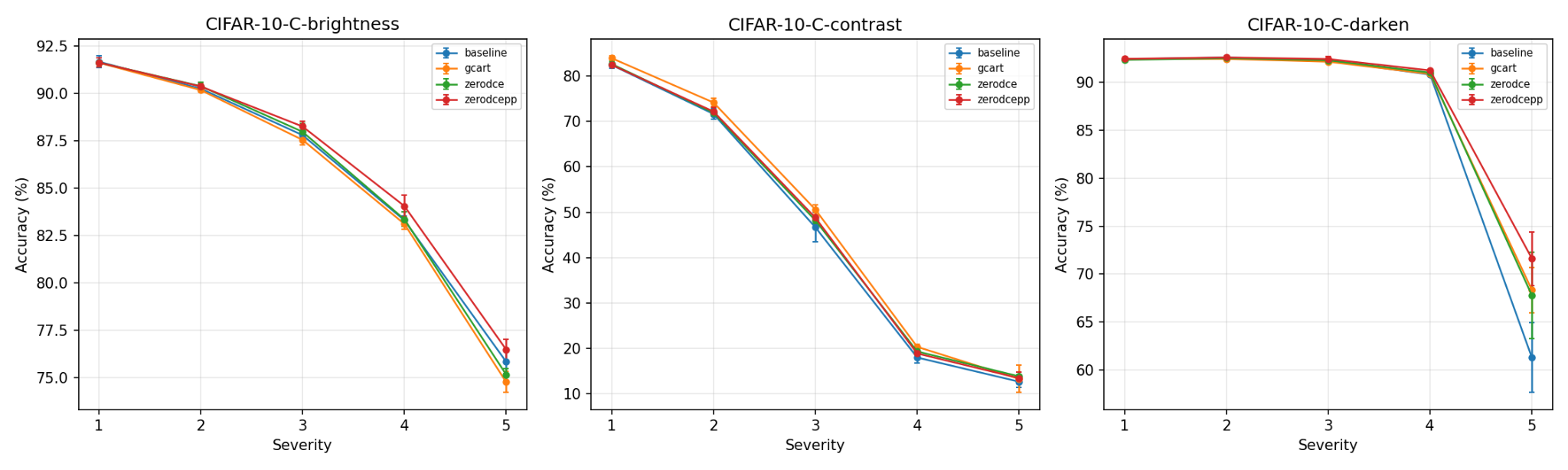}
\caption{Accuracy as a function of corruption severity for the four learned systems on CIFAR-10-C-style brightness, contrast, and darkening corruptions. Mean and standard deviation over 3 seeds.}
\label{fig:severity}
\end{figure*}

Table~\ref{tab:main} reports clean accuracy and the severity-averaged accuracy on each corruption for the four learned systems. Figure~\ref{fig:severity} shows the same numbers resolved by severity.

\begin{table}[t]
\centering
\small
\setlength{\tabcolsep}{4pt}
\begin{tabular}{@{}lcccc@{}}
\toprule
\textbf{Model} & \textbf{Clean} & \textbf{Bright.} & \textbf{Contrast} & \textbf{Darken} \\
\midrule
Baseline    & 92.23$\pm$0.09 & 85.77$\pm$0.09 & 46.27$\pm$0.89 & 85.87$\pm$0.63 \\
Zero-DCE    & 92.23$\pm$0.12 & 85.69$\pm$0.18 & 47.17$\pm$0.47 & 87.18$\pm$0.78 \\
Zero-DCE++  & 92.31$\pm$0.19 & \textbf{86.15}$\pm$0.28 & 47.13$\pm$0.71 & \textbf{88.06}$\pm$0.38 \\
GC-ART      & 92.12$\pm$0.15 & 85.44$\pm$0.18 & \textbf{48.45}$\pm$0.27 & 87.24$\pm$0.34 \\
\bottomrule
\end{tabular}
\caption{Clean-test and corruption-averaged accuracy (mean $\pm$ s.d.\ over 3 seeds, all values percent). Corruption columns are averaged over severities 1--5. Best per column in bold, but note that all corruption-column gaps between learned systems are within or near 1$\sigma$ of seed variance.}
\label{tab:main}
\end{table}

Table~\ref{tab:main} shows that all learned systems preserve high clean accuracy: the clean means fall within 0.19 percentage points of each other. On \textsc{brightness}, the learned systems remain close, with a 0.71~pp spread between the best and worst means. On \textsc{darken}, all three learned enhancers improve over the baseline by 1.3--2.2~pp on average, showing that a learned front-end can provide useful underexposure correction beyond the clean classifier. GC-ART is part of this improvement, and therefore the darkening result is a positive outcome even though the three learned enhancers are close to one another. On \textsc{contrast}, GC-ART obtains the highest learned-method average (48.45\% vs. 46.27\% for the baseline and 47.13\% for Zero-DCE++). This is the most diagnostic corruption for GC-ART's design because recovering contrast requires expanding compressed intensity ranges, rather than applying a one-directional brightening or darkening correction.

Overall, the learned-front-end results highlight GC-ART's contrast behavior. GC-ART outperforms the baseline, MiniZeroDCE, and MiniZeroDCE++ on the severity-averaged contrast metric while matching their clean accuracy. Because the contrast corruption collapses intensities toward the per-channel mean, undoing it requires a bidirectional remapping that pushes low and high intensities apart. This result is evidence that a global rational curve conditioned on image histograms can learn useful contrast-expansion behavior. The effect is still much smaller than the gain from classical HE on this synthetic corruption, so we treat it as a promising direction rather than a complete robustness solution.

\subsection{Comparison Against Classical Pre-processors}
\label{sec:classical}
A more striking comparison is against classical, parameter-free pre-processors --- Histogram Equalization (HE), CLAHE, and three gamma-correction settings --- applied at both training and test time. Table~\ref{tab:classical} reports these results in the same format as Table~\ref{tab:main}.

\begin{table}[t]
\centering
\small
\setlength{\tabcolsep}{4pt}
\begin{tabular}{@{}lcccc@{}}
\toprule
\textbf{Pre-proc.} & \textbf{Clean} & \textbf{Bright.} & \textbf{Contrast} & \textbf{Darken} \\
\midrule
HE                & 89.81$\pm$0.18 & 82.00$\pm$0.39 & \textbf{89.12}$\pm$0.22 & \textbf{89.72}$\pm$0.22 \\
CLAHE             & 89.99$\pm$0.13 & 75.10$\pm$0.97 & 58.67$\pm$0.68 & 82.94$\pm$0.74 \\
$\gamma{=}1.5$    & 92.21$\pm$0.06 & 85.66$\pm$0.08 & 47.19$\pm$0.50 & 88.36$\pm$0.37 \\
$\gamma{=}2.2$    & 92.24$\pm$0.13 & 85.78$\pm$0.26 & 47.68$\pm$0.42 & 88.66$\pm$0.47 \\
$\gamma{=}3.0$    & \textbf{92.35}$\pm$0.12 & \textbf{85.92}$\pm$0.08 & 48.01$\pm$0.12 & 87.88$\pm$0.58 \\
\bottomrule
\end{tabular}
\caption{Classical pre-processors applied at both training and test time (mean $\pm$ s.d., 3 seeds, percent). Best per column in bold; HE dominates two of three corruption columns by large margins.}
\label{tab:classical}
\end{table}

Table~\ref{tab:classical} provides a useful calibration for the learned-enhancer results. Histogram Equalization is very strong on the two corruptions that are closest to monotone global intensity transforms: it reaches 89.12\% on \textsc{contrast} and 89.72\% on \textsc{darken}. At the highest contrast severity, HE retains 87.14\% accuracy.

This behavior is expected for the synthetic corruptions used here. HE maps each image's intensities toward an approximately uniform distribution, making the processed representation relatively insensitive to monotone global contrast changes. The tradeoff is lower clean accuracy: HE loses roughly 2.4~pp relative to the learned systems. Gamma correction also remains competitive. At $\gamma=2.2$, it matches or exceeds every learned enhancer on three of the four reported columns while adding no learnable parameters. CLAHE, in contrast, underperforms plain HE on every column, suggesting that local equalization is not helpful for these global corruption transforms.

For CIFAR-10 at $32\times32$ under these illumination corruptions, HE and fixed gamma correction are strong baselines that any learned tone-mapping front-end should be compared against. GC-ART's contribution is therefore complementary: it offers a compact, trainable, edge-preserving tone curve that matches clean accuracy and shows the best learned-method contrast result, while classical pre-processing defines a high target for future learned modules.

\subsection{Computational Cost: FLOPs}
\label{sec:cost}

\begin{table}[t]
\centering
\small
\setlength{\tabcolsep}{5pt}
\begin{tabular}{@{}lrr@{}}
\toprule
\textbf{Module} & \textbf{Params} & \textbf{Total FLOPs (32)} \\
\midrule
GC-ART                &     643 &     269{,}088 \\
Zero-DCE              &  11{,}011 &  11{,}252{,}736 \\
Zero-DCE++            &   1{,}953 &   1{,}908{,}736 \\
Histogram Equaliz.    &       0 &      19{,}200 \\
Gamma ($\gamma{=}2.2$)&       0 &       6{,}144 \\
\bottomrule
\end{tabular}
\caption{Parameter counts and FLOPs at $32{\times}32$. \emph{Total FLOPs} includes parameter-prediction MACs and pixel-operation FLOPs.}
\label{tab:cost}
\end{table}

Table~\ref{tab:cost} reports parameter counts and FLOPs at $32\times32$. GC-ART is substantially cheaper than the convolutional learned enhancers by this measure: GC-ART uses 269k FLOPs, compared with 11.25M for the Zero-DCE-style convolutional enhancer and 1.91M for Zero-DCE++. This corresponds to roughly 42$\times$ fewer FLOPs than Zero-DCE and roughly 7$\times$ fewer FLOPs than Zero-DCE++ at CIFAR resolution.

The FLOP comparison reflects the different scaling structure of the modules. Zero-DCE-style convolutional MACs scale with $H\cdot W$ across multiple learned spatial filters. In GC-ART, the learned MLP that predicts the curve parameters is independent of resolution, while the soft histogram and pointwise curve application scale with the number of pixels. Thus the accurate cost claim is not that GC-ART has constant total enhancement cost, but that its learned parameter-prediction component is resolution-independent and its total FLOP count is much lower than the convolutional learned enhancers in this setup.

The classical pre-processors remain the cheapest methods in FLOPs. Gamma correction is a single pointwise operation, and Histogram Equalization has no learned parameters and lower counted FLOPs than GC-ART in this benchmark. These results make classical methods strong computational baselines. GC-ART occupies a different point in the design space: it is trainable and image-adaptive, uses a very small learned parameter-prediction network, and avoids spatial filtering while remaining far smaller in FLOPs than the convolutional learned enhancers.


\section{Conclusion}
We introduced GC-ART, a histogram-conditioned rational tone-mapping module with a 643-parameter MLP, and evaluated it as a learned front-end for CIFAR-10 classification under illumination corruptions. GC-ART matches the clean accuracy of the unenhanced classifier and Zero-DCE-style learned enhancers, improves over the unenhanced baseline on darkening, and achieves the strongest learned-method result on contrast corruption. The darkening result shows useful learned underexposure correction, while the contrast result supports the central motivation of the module: a global rational curve conditioned on image statistics can learn bidirectional contrast-expansion behavior, pushing compressed dark and bright ranges apart within the same image. Its pointwise form also preserves spatial edge locations by construction, unlike local image-to-image enhancement networks that may introduce smoothing or local artifacts. Classical Histogram Equalization remains a very strong reference on these synthetic global corruptions, though its drawbacks are discussed in Section 3.4. The FLOP analysis shows that GC-ART is much cheaper than the convolutional learned enhancers by counted operations, while classical HE and gamma correction remain cheaper fixed pre-processing baselines. The GC-ART hyperparameters used here, including histogram bin count, bandwidth, MLP width, monotonicity weight, and monotonicity grid size, were not systematically tuned; optimizing these choices may further improve performance. Overall, GC-ART provides a compact, adaptive, and edge-preserving tone-mapping design, with the present contrast results motivating more targeted study of learned global curves for robust recognition.

\section{Scope and Limitations}
\label{sec:limitations}
We summarize the experimental scope and the main caveats for interpreting the results.

\textbf{Dataset and corruption scope.}
All accuracy numbers are for CIFAR-10 at $32{\times}32$ with ResNet-18 and three illumination corruptions implemented as global pixel transforms. We do not test natural OOD imagery, higher resolution, other backbones, other classification tasks, or non-illumination corruptions (noise, blur, weather, etc.). Findings about the relative ranking of methods, and in particular HE's strong performance on contrast, may not generalize outside this setup.

\textbf{Synthetic, monotone illumination corruptions.}
The brightness, contrast, and darken transforms we use are global, monotone pixel-wise functions, applied uniformly across each image. HE's invariance to monotone intensity transforms is exactly the structure that produces its strong performance on contrast and darkening. On natural night-time imagery, where corruptions are spatially varying (cast shadows, hot spots, sensor noise), this advantage may not transfer directly. The reader should interpret the HE result as showing that on \emph{this particular corruption family} a global histogram-uniformizing operation is a strong reference point for learned tone maps, rather than as a complete deployment conclusion.

\textbf{Limited number of seeds.}
Our seed-variance estimates are over $n=3$ runs each. The standard deviations in Tables~\ref{tab:main} and~\ref{tab:classical} should therefore be read as rough indicators rather than tight confidence intervals, and small ranking differences within the learned-method group should not be over-interpreted.

\textbf{Simplified Zero-DCE-style baselines.}
The systems labeled ``Zero-DCE'' and ``Zero-DCE++'' are stripped-down variants trained with classification cross-entropy only, sharing the LE curve formulation with the original Zero-DCE~\cite{guo2020zerodce} but omitting four reference losses and using fewer layers and iterations. These comparisons do not evaluate Zero-DCE as a full method or measure the value of its perceptual reference losses; they only characterize this architectural family under classification-only training.

\textbf{Bidirectional capacity needs more direct analysis.}
The motivation for the rational curve form (Section~\ref{sec:curve}) appeals to the family's ability to express both shadow-lifting (concave) and highlight-compressing (convex) shapes. Our corruption set gives partial evidence for this motivation. \textsc{darken} and \textsc{brightness} test separate under- and overexposure axes, while \textsc{contrast} collapse requires a bidirectional correction within a single image: low and high intensities must be pushed apart around the mean. GC-ART is the best learned method on the contrast metric, which supports the value of rational bidirectional tone shaping in this setting. A more direct test would analyze the predicted curves for normal, contrast-compressed, brightened, and darkened inputs and verify that the learned shapes implement the intended correction mechanisms. This analysis is the first item of Section~\ref{sec:future}.

\textbf{Output range.}
The rational curve has $f(0)=0$ and $f(1)=1$ by construction, but $f(x)$ for intermediate $x$ can fall outside $[0,1]$ for some parameter values; the current implementation does not clamp the output. This did not destabilize training in our runs. Future versions could add a range-preserving parameterization or an explicit clamp if strict image bounds are desired.

\textbf{Monotonicity regularization.}
$\lambda{=}10$ was chosen by hand as the weight on $\mathcal{L}_{\text{mono}}$, which itself is reduced by mean over batch, channel, and finite-difference sample axes. We did not sweep $\lambda$, so the sensitivity of GC-ART to this regularizer remains to be characterized. Monotonicity may become more important on harder corruption regimes or higher-resolution data.

\textbf{Run-to-run determinism.}
The training script enables cuDNN benchmark mode and \texttt{torch.compile} autotuning. Both perform autotuning that can select different kernels across runs, so individual seed accuracies are not bit-reproducible even on the same hardware; this is part of what the seed-variance numbers reflect.


\section{Future Directions}
\label{sec:future}
Several extensions would clarify the role of GC-ART and learned tone mapping more generally.

\textbf{Hyperparameter sweeps.}
The present experiments fix several module choices by hand: MLP hidden width 32, monotonicity weight $\lambda=10$, histogram bin count $K=16$, bandwidth $\gamma=0.01$, and monotonicity grid size $M=32$. A systematic sweep over width, $\lambda$, $K$, and $\gamma$ would determine how much additional performance is available from the same basic design.

\textbf{Curve behavior analysis.}
The rational family can represent both shadow-lifting and highlight-compressing curves, and the contrast results suggest that this flexibility may help when compressed intensity ranges must be expanded. Plotting the predicted curves for normal, brightened, darkened, and contrast-compressed inputs would reveal whether the model learns distinct correction modes or collapses near identity. This analysis should include the slope of the learned curve in shadow, mid-tone, and highlight regions, since contrast recovery should correspond to steeper slopes over the compressed intensity range. It would also indicate whether larger MLPs, weaker monotonicity regularization, or richer histograms, such as joint RGB histograms, are likely to help.

\textbf{Original Zero-DCE comparison.}
Our Zero-DCE-style baselines are simplified and trained only with classification loss. A complementary experiment would evaluate the original Zero-DCE network trained with its perceptual zero-reference losses, either frozen as a front-end or fine-tuned jointly with the classifier. This would separate the spatial-versus-global architectural question from the effect of the original enhancement objective.

\textbf{Edge sharpness and local artifacts.}
Because GC-ART applies a global pointwise curve, it preserves the spatial support of edges: a boundary may change in contrast, but the module cannot mix pixels across the boundary. This is a potential advantage over sliding-window CNN or image-to-image enhancement front-ends, which can blur edges or introduce local artifacts if their learned filters over-smooth. The present paper does not quantify this property. Future work should measure edge sharpness, boundary localization, and downstream recognition on high-resolution images where local texture and boundary preservation matter.

\textbf{Broader data and corruptions.}
All experiments here use CIFAR-10, $32\times32$ resolution, ResNet-18, and synthetic global illumination corruptions. Higher-resolution datasets, natural low-light or HDR imagery, other backbones, and non-illumination corruptions such as noise, blur, and weather may change the relative behavior of learned and classical pre-processing. These settings are also more likely to expose the edge-preservation advantage of pointwise global curves, since CIFAR-10 is too low-resolution to evaluate perceptual sharpness reliably.

\section{Reproducibility}
\label{sec:repro}
The full training code, the CIFAR-10 wrapper, model definitions, evaluation pipeline, classical-baseline pre-processors, aggregation scripts, and FLOPs benchmark script are available at \texttt{https://github.com/joyc2point718/gcart}. The training scripts write per-seed JSON output files and the aggregation script writes the derived summary tables under \texttt{results/}; these raw outputs should be included with the accompanying artifact for exact reproduction of the reported numbers.


\end{document}